\documentclass[journal]{IEEEtran}
\usepackage{amsmath,amsfonts}
\usepackage{algorithmic}
\usepackage{algorithm}
\usepackage{array}
\usepackage[caption=false,font=normalsize,labelfont=sf,textfont=sf]{subfig}
\usepackage{textcomp}
\usepackage{stfloats}
\usepackage{url}
\usepackage{verbatim}
\usepackage{graphicx}
\usepackage{cite}
\usepackage{color}
\usepackage{fancyhdr}

\usepackage{multirow}
\usepackage{color}
\usepackage[driverfallback=dvipdfm,
CJKbookmarks=true, bookmarksnumbered=true, bookmarksopen=true,
colorlinks=true,
citecolor=blue, linkcolor=blue, anchorcolor=red, urlcolor=black
]{hyperref}
\hypersetup{hidelinks,	
            colorlinks=true,	
            allcolors=black,	
            pdfstartview=Fit,	
            breaklinks=true}

\begin{document}

\title{Hierarchical Forgery Classifier \\On Multi-modality Face Forgery Clues}

\author{Decheng~Liu,~Zeyang~Zheng,~Chunlei~Peng,~\IEEEmembership{Member, IEEE},~Yukai~Wang,~Nannan~Wang,~\IEEEmembership{Member, IEEE}~and~Xinbo~Gao,~\IEEEmembership{Senior Member, IEEE}        
}

\markboth{~Vol.~xx, No.~xx, August~2023}%
{IEEE Transactions on Multimedia}


\maketitle
\thispagestyle{fancy}
\cfoot{\small{\copyright~2023 IEEE. Personal use of the material is permitted. 
Permission from IEEE must be obtained for all other uses, in any current or future media, including reprinting/republishing this material for advertising or promotional purposes, creating new collective works, for resale or redistribution to servers or lists, or reuse of any copyrighted component of this work in other works.}}
\rfoot{}

\begin{abstract}
Face forgery detection plays an important role in personal privacy and social security. With the development of adversarial generative models, high-quality forgery images become more and more indistinguishable from real to humans. 
Existing methods always regard as forgery detection task as the common binary or multi-label classification, and ignore exploring diverse multi-modality forgery image types, e.g. visible light spectrum and near-infrared scenarios.
In this paper, we propose a novel \textbf{H}ierarchical \textbf{F}orgery \textbf{C}lassifier for \textbf{M}ulti-modality \textbf{F}ace \textbf{F}orgery \textbf{D}etection \textbf{(HFC-MFFD)}, which could effectively learn robust patches-based hybrid domain representation to enhance forgery authentication in multiple modality scenarios.
The local hybrid domain representation is designed to explore strong discriminative forgery clues both in the image and frequency domain with the intra-attention mechanism.  
Furthermore, the specific hierarchical face forgery classifier is designed through the authenticity feedback strategy to integrate diverse discriminative clues.
Experimental results on representative multi-modality face forgery datasets demonstrate the superior performance of the proposed HFC-MFFD compared with state-of-the-art algorithms.
\emph{The source code and models are publicly available at \href{https://github.com/EdWhites/HFC-MFFD}{https://github.com/EdWhites/HFC-MFFD}.}

\end{abstract}


\section{Introduction} \label{section:Introduction}
\begin{figure}
    \centering
    \includegraphics[width=0.5\textwidth]{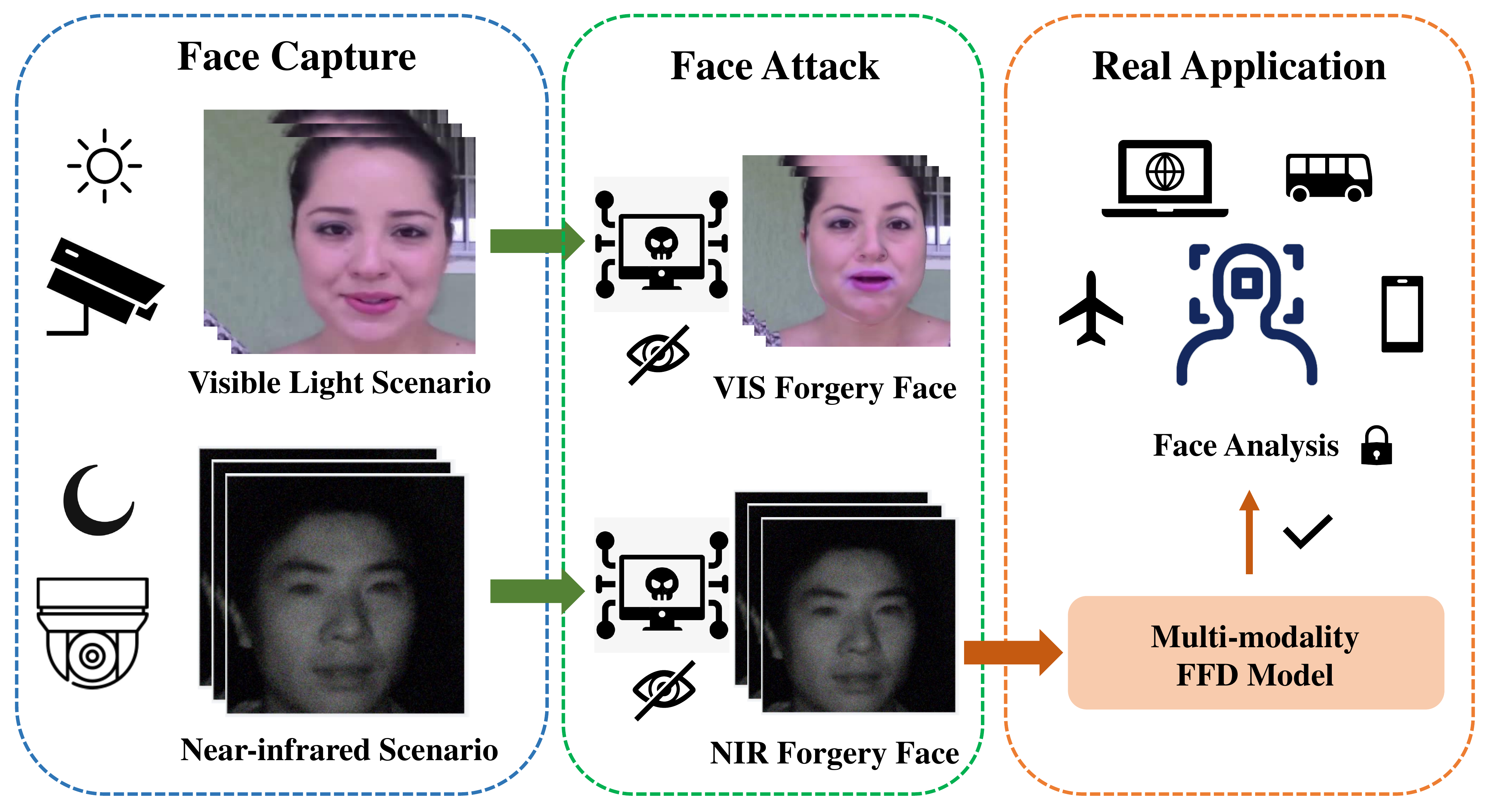}
    \caption{
    The diagram of multi-modality face forgery detection task in the real-world application.}
    \label{fig:application}
\end{figure}

\IEEEPARstart{R}{ecently} researchers have developed the deep generative model rapidly, 
and face manipulation technology also has significant growth \cite{guera2018deepfake,li2020celeb,karaouglu2021self,song2021agegan++,he2021forgerynet}.
These high-quality face forgery images or videos even become indistinguishable to human eyes, which inevitably brings negative effects on social security.
The forgery technology could be used to create fake news and spread misleading information.
Thus, face forgery detection algorithm has drawn more and more attention in the computer vision field.
Most related works regard forgery detection as a simple binary classification task, and great progress has been achieved with deep learning development.
In the real world, face forgery images could be generated by diverse generative models and captured in variant illumination.
Thus, face forgery detection is still a challenging problem, especially for multi-modality data.
As shown in Fig. \ref{fig:application}, different modalities of images are captured by diverse types of cameras in both visible light and near-infrared scenarios.
To effectively distinguish fake images when encountering attack behaviors, the multi-modality face forgery detection (MFFD) model plays a necessary role before following face analysis tasks (e.g. expression, attributes and identity recognition) in the real world.

Existing face forgery detection methods could be roughly classified into image domain-based and frequency domain-based detection methods.
Early studies always designed a specific image classification model to identify real or fake \cite{wang2022forgerynir,chollet2017xception,yu2019attributing,huang2021initiative}.
These methods only focus on spatial information clues in face forgery and ignore other forgery clues.
\cite{frank2020leveraging} firstly find that there indeed exist obvious grid-like visual artifacts in the frequency spectrum in fake images.
Inspired by the phenomenon, many researchers have proposed diverse forgery clues discovery methods through the frequency domain information \cite{qian2020thinking,liu2021spatial,li2021frequency}.
However, previously mentioned works focus on exploring the forgery clues in the visible light image (VIS), and lack of exploration in the near-infrared images (NIR).
Researchers\cite{wang2022forgerynir} find previous forgery detection for VIS faces methods perform poorly in NIR face forgery detection tasks.
It is because of the complex modality gap and overfitting in the deep learning model.

\begin{figure*}[ht]
    \centering
    \includegraphics[width=1\textwidth]{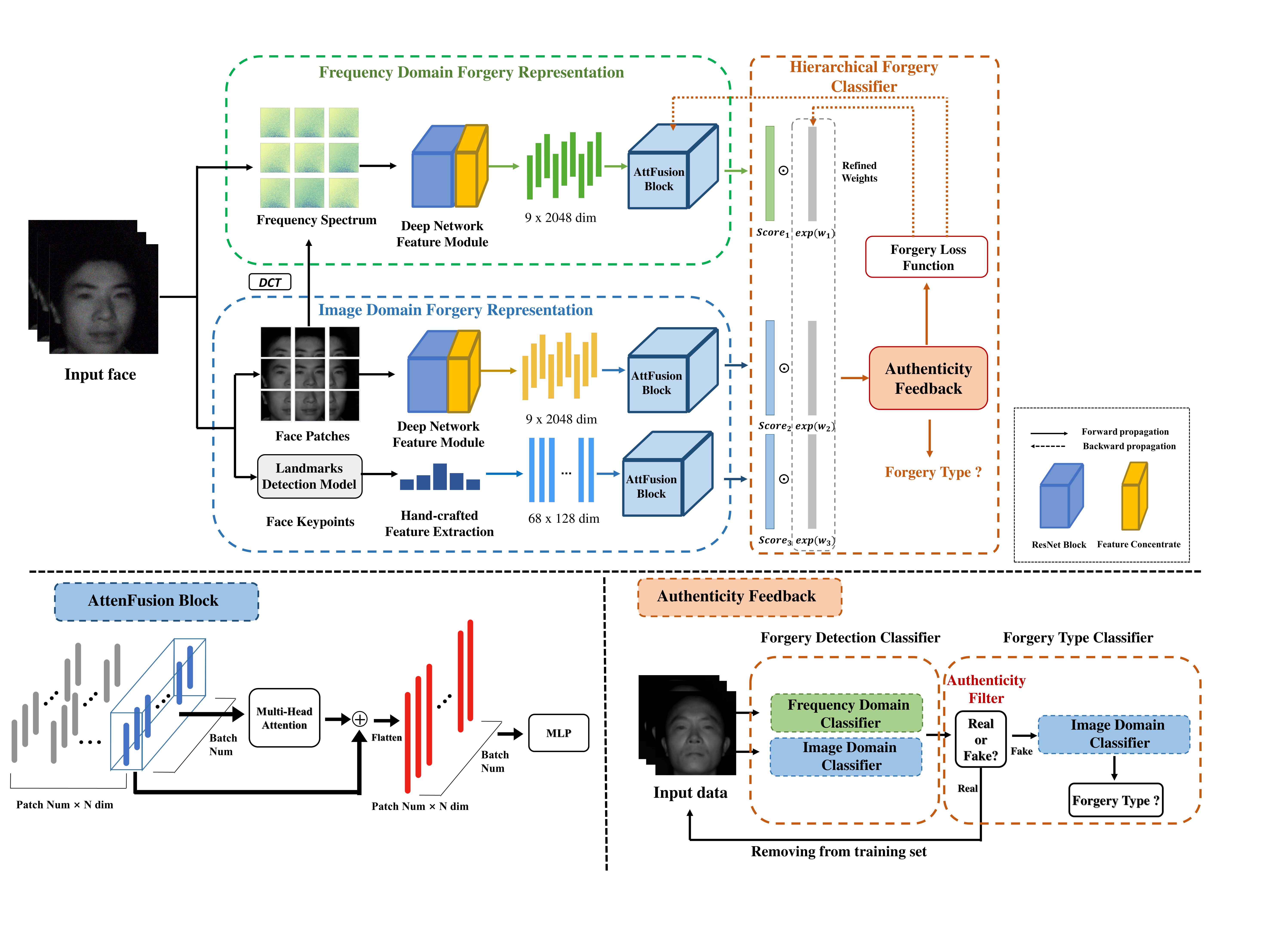}
    \caption{
    The overview of the proposed hierarchical forgery classifier on multi-modality face forgery clues. The left-bottom subfigure shows the architecture of the designed Attention Fusion (AttenFusion) Block, and the right-bottom subfigure shows the detailed architecture of the designed Authenticity Feedback Module.
    }
    \label{fig:framework}
\end{figure*}

To solve mentioned problems, this paper proposes a novel hierarchical forgery classifier on multi-modality face forgery clues, which could effectively learn robust path-based hybrid domain representation and further design the specific hierarchical forgery classifier to identify fake images and artifact types.
The framework of the proposed algorithm is shown in Fig. \ref{fig:framework}.
Inspired by intrinsic characteristics \cite{Neucom2018,LiuPR21} in multi-modality face images, we first divide the holistic faces into local patches to make features more robust in complex scenarios. 
Related works \cite{chai2020makes,chen2021local} also prove that the local patches but not the global facial structure, are more likely stereotyped and contain redundant artifact clues.
Considering the limited number of training data, we extract both hand-crafted features and deep network features for easing overfitting problems.
Because of complementary discriminative forgery information extracted in mentioned features, we further utilized the score fusion strategy to enhance the detection performance.
It is noted that the number of real images is always much less than the number of fake images, which could be generated by various image generation pipelines.
Different from \cite{wang2022forgerynir} regarding the task as the five classes classification task, we newly design a two-stage hierarchical forgery classifier, which could reduce the adverse effects of the class imbalance problem. 
The binary forgery detection classifier aims to distinguish the authenticity of the input face images.
Here the proposed hybrid domain discriminative representation could learn in both image patches and frequency spectrum.
Finally, the fake images are input into the forgery type classifier to recognize the artifact types.
The unique aspect of the proposed algorithm lies on: (1) the hierarchical forgery classifier utilizing the authenticity feedback strategy for multi-modality faces; (2) the local hybrid domain forgery representation consisting the attention fusion block to learn dependencies between different spatial patches to explore robust forgery clues.
\emph{The main reason for superior performance is that we effectively design patch-based hybrid forgery representation to improve the generalization, and the hierarchical classifier pipeline to enhance performance.
}

\IEEEpubidadjcol
The main contributions of our paper are summarized as follows:
\begin{enumerate}
\item We employ the hierarchical forgery classifier for the multi-modality face forgery analysis task. It utilized the authenticity feedback strategy to distinguish the authenticity and recognize artifact types in order, which could integrate diverse discriminative clues to boost performance.
\item The local hybrid domain forgery representation is introduced in both image and frequency domains. The designed attention fusion block effectively learns the long-range dependencies between different patches, which can explore robust discriminative forgery clues for multi-modality images.
\item Experimental results illustrate the superior performance of the proposed method compared with state-of-the-art forgery detection algorithms in multi-modality face forgery datasets.
\emph{The code is publicly available at \href{https://github.com/EdWhites/HFC-MFFD}{https://github.com/EdWhites/HFC-MFFD}}.

\end{enumerate}

We organized the rest of this paper as follows. Section \ref{section:Introduction} gives a brief introduction of the proposed method, and section \ref{section:Related work} shows some representative face forgery detection algorithms. In Section \ref{section:Proposed method}, we present the novel hierarchical forgery classifier on multi-modality face forgery clues. Section \ref{section:Experiments} shows the experimental results and analysis. The conclusion is drawn in Section \ref{section:Conclusion}.

\section{RELATED WORK} \label{section:Related work}
Face forgery detection has been studied as an important problem for many years. Here we review the representative forgery detection methods in two categories: image domain-based forgery detection methods and frequency domain-based forgery methods.
Heterogeneous face analysis-related works are summarized following.

\subsection{Image Domain Forgery Detection}

Early approaches for forgery detection are performed by evaluating hand-crafted features from algorithms such as local binary patterns \cite{ojala2002multiresolution}, scale-invariant feature transform \cite{lowe2004distinctive} and histogram of Oriented Gradient \cite{dalal2005histograms} with classifier like support vector machine \cite{cortes1995support}. In addition to giving classification results directly, SVM can also output scores for binary classification\cite{platt1999probabilistic} or multi-class classification\cite{wu2003probability}. And in recent years, with the development of deep learning, especially after the proposal of VGGNet \cite{simonyan2014very} and ResNet \cite{he2016deep}, the number of methods based on convolutional neural networks \cite{mo2018fake,wang2020cnn,liu2023fedforgery} for forgery detection is increasing, and most of them have achieved impressive results. For instance, Li \emph{et al.} \cite{li2020face} proposed a method that leverages data from a richer and related domain to learn meaningful features in the application-specific domain for which training samples are scarce through the concept of neural network distilling. 
Considering that the Transposed Convolution in GAN will cause the lack of global information in the generated images, Mi \emph{et al.} \cite{mi2020gan} proposed a method using self-attention to grasp the global information of the input image after local features extraction of the convolution layer. 
Fei \emph{et al.} \cite{fei2020attribute} proposed a method that turns the forgery detection task into a restoration task. After the selected attributes of the image are erased, the error between the restored image and the original image will be measured which will be then used for forgery detection.
To reduce the time cost for designing effective DCNNs for specific image forensics tasks, Chen \emph{et al.} \cite{chen2020automated} design an automated network trained through Q-learning with the epsilon-greedy strategy and experience replay.

Face forgery video detection is another important topic.  Apart from converting the video into the frameset at the image level, some studies also study the anomalies of the forgery videos at the video level directly. 
Su \emph{et al.}\cite{su2017fast} utilized the ExponentialFourier moments to design a fast forgery detection algorithm to detect region duplication.
Chen \emph{et al.}\cite{chen2015automatic} proposed a method that utilizes motion residuals as input, with image steganalysis methods to extract features, and then be classified by random forest. 
LIY \emph{et al.} \cite{liy2018exposingaicreated} proposed a novel method that performs forgery detection on videos by detecting eye blinking. 
Peng \emph{et al.} \cite{peng2020temporal} designed a network to perform forgery detection by analyzing temporal consistency in videos. Chintha \emph{et al.} \cite{chintha2020recurrent} proposed the XcepTemporal network, which consists of XceptionNet for face image feature extraction and LSTM for temporal inconsistency detection. This network can detect both spatial and temporal signatures of deepfake renditions after extracting the primary face from each frame in the video. 
Yang \emph{et al.}\cite{yang2020efficient} proposed a video-container-based method to detect forgery videos, which is based on useful forensic traces in the video container structure.
For copy-move forgeries, D’Amiano \emph{et al.} \cite{d2018patchmatch} proposed a method that computes features invariant to various spatial, temporal, and intensity transformations, and then build a nearest-neighbor field for matching, with a series of criterion to tell apart copy moves from false matches.
Aloraini \emph{et al.} \cite{aloraini2020sequential} proposed two methods named sequential analysis and patch analysis which model video sequence in different ways, with the aim to detect a video forgery by capturing the changes in the parameters or identifying the distribution, after applying a spatiotemporal filter to capture forgery traces.
Besides, the forensics of compressed videos is still challenging work. 
Hu \emph{et al.} \cite{hu2021detecting} proposed a method using a frame-level and temporality-level stream to extract two kinds of features before inputting them into two different convolutional neural networks to obtain scores. The network in the frame-level stream will be gradually pruned to avoid being influenced by the redundant artifacts noise. He \emph{et al.} \cite{he2019exposing} found that the recompression error generated by the recompression operation has a large difference in the spatial domain statistics between the real video and the forged video, which could be regarded as the forgery clue.
Zhang \emph{et al.} \cite{zhang2022unsupervised} firstly proposed a novel unsupervised forgery detection method to distinguish fake videos according to different provenance properties.
Song \emph{et al.} \cite{song2022face} proposed a symmetric transformer to eliminate the prediction instability in the forgery detection task, which utilized the channel variance and spatial gradients as the feature.
Nowadays, most related works still regard forgery detection as a specific image classification task, ignoring the instinctive property of the real-world task.

\subsection{Frequency Domain Forgery Detection} 

Recently related works begin to pay attention to the abnormal situation that happened in the frequency domain of forgery images. 
Zhang \emph{et al.} \cite{zhang2019detecting} showed theoretically that checkboard artifacts are manifested in the DFT spectrum. Dzanic \emph{et al.} \cite{dzanic2020fourier} proposed to use the difference in the decay rate of the high-frequency spectrum between the forgery and the real images for forgery detection while Chandrasegaran \emph{et al.} \cite{chandrasegaran2021closer} showed that this discrepancy can be avoided by change the last upsampling operation. Frank \emph{et al.} \cite{frank2020leveraging} observed that upsampling operations would cause grid-like patterns in the DCT spectrum which can be linearly separated. Giudice \emph{et al.} \cite{giudice2021fighting} did further research, dividing an image into patches with a size of 8$\times$8 and then transforming them into the DCT spectrum for forgery detection. Inspired by the observation that frequency information could be a clue, Qian \emph{et al.} \cite{qian2020thinking} designed the F$^{3}-$Net consisting of two frequency-aware branches performing forgery detection through frequency-aware image decomposition and forgery clues exploitation in the raw frequency domain.

Considering the complementary information in the image domain and the frequency domain, fusing them from two different domains might be helpful for boosting performance. Song \emph{et al.} \cite{song2022adaptive} explored mining the potential consistency through the correlated representations from RGB and frequency domains to conduct face forgery detection.
It also inspired recent works\cite{liu2021spatial,li2021frequency,wang2022m2tr} to exploit the forgery clues in both the image domain and frequency domain.
However, how to effectively integrate two different complementary discriminative information in multi-modality images is still a challenging problem.

\subsection{Multi-modality Face Analysis}
The goal of multi-modality face analysis is to analyze input face images with important semantic information (like attributes and identity).
Existing approaches can be generally grouped into three categories: {synthesis-based} methods, common space projection based methods, and feature descriptor based methods.
(1) {Synthesis-based} methods first transform the heterogeneous face images into the same modality. Once the synthesized photos are generated from non-photograph images or vice versa, conventional face recognition algorithms can be applied directly. 
The global eigen transformation was proposed \cite{10} to synthesize face sketches from face photos. 
\cite{11} presented a patch-based approach with the idea of locally linear approximating global nonlinear, which represents each target sketch patch by a linear combination of some candidate sketch patches. 
Sparse representation \cite{39}was also applied to face sketch synthesis to compute the combination weight. 
\cite{5} adopted a fully convolutional network (FCN) to directly model the complex nonlinear mapping between face photos and sketches. 
Liu \textit{et al.} \cite{LiuPR21} proposed the novel iterative local re-ranking algorithm to process diverse synthesis faces, which is generated by different attribute clues.
However, the synthesis process is actually more difficult than recognition and the performance of these methods heavily depends on the fidelity of the synthesized images.
(2) Common space projection based methods attempt to project face images in different modalities into a common subspace where the discrepancy is minimized. Then heterogeneous face images can be matched directly in this common subspace. 
Kan \textit{et al.} \cite{Ref11} proposed a multi-view discriminant analysis (MvDA) method to obtain a discriminant common space for recognition. The correlations from both inter-view and intra-view were exploited.
Sharma and Jacobs \cite{sharma2011bypassing} proposed the partial least squares algorithm to learn the linear mapping between different face modalities.
Huo \textit{et al} \cite{TCyMargin} proposed a margin-based cross-modality metric learning to address the gap between different modalities.
Liu \textit{et al} \cite{TNNLS20} directly utilized the deep learning model as the mapping function, which is also integrated with extra semantic attribute information.
Hu \textit{et al} \cite{hu2020adversarial} proposed a method that uses modality-adversarial feature learning and attention network to reduce the gap of cross-modality images.
Liu \textit{et al} \cite{liu2021heterogeneous} leveraged the interpretable disentangled face representation to conduct both synthesis and recognition tasks.
However, the projection procedure always causes information loss which decreases the recognition performance.
(3) Feature descriptor based methods first represent face images with local feature descriptors. These encoded descriptors can then be utilized for recognition. 
Lei \textit{et al.} \cite{Ref25} proposed a discriminant image filter learning method that benefitted from LBP-like face representation for matching NIR to VIS face images. 
Alex \textit{et al.} \cite{Ref23} proposed a local difference of Gaussian binary pattern (LDoGBP) for face recognition across modalities.
Han \textit{et al.} \cite{Ref24} proposed a component-based approach for matching composite sketches to mug shot photos.
Liu \textit{et al.} \cite{Neucom2018} further fuse different face components discriminative information to boost recognition performance.
Nowadays, feature descriptor based methods always achieve better recognition performance.

Nowadays, most forgery detection works are deployed on VIS images, experimental results prove the poor generalization ability in NIR scenarios.
With the development of NIR cameras in the real world, researchers need to pay more attention to multiple modality face forgery detection tasks.

\section{PROPOSED METHOD} \label{section:Proposed method}

In this section, we present a face forgery classification algorithm, which utilizes the designed Local Hybrid Domain Forgery Representation. 
It is noted that we adopt a Hierarchical Face Forgery Classifier for forgery detection and forgery type tracing respectively. 
Fig. \ref{fig:framework} shows the framework of the proposed algorithm. 
In the subsection, we will describe the detailed explanation of the algorithm, which consists of three parts: Preprocessing, Local Hybrid Domain Forgery Representation and Hierarchical Face Forgery Classifier.


\subsection{Local Hybrid Domain Forgery Representation}
The input image is divided into nine patches in both image and frequency domains with the designed preprocessing procedures.
Inspired by former heterogeneous face analysis works \cite{Neucom2018, LiuPR21}, we separately design hand-crafted features and deep network features to extract discriminative complementary information to construct local hybrid domain forgery representation for multiple modality face images.
The details are given below.


Firstly, we aim to explore the multi-modality forgery clues in the image domain.
Inspire by related works \cite{chai2020makes,chen2021local}, we find local patches of input faces are more likely to contain redundant artifacts.
Thus, we utilize different feature extraction methods to learn suitable effective features in the global and local spatial relationship.
For the local patches of generated faces, we select ResNet as the deep neural network to extract deep network features for high-level semantic discriminative information.
Additionally, we select the SIFT algorithm as the typical hand-crafted feature to capture more texture discriminative information. It is because the hand-craft feature could maintain good discriminability even under disturbance attacks. Here we utilize the existing face landmark detector algorithm \footnote{http://dlib.net/files/shape\_predictor\_68\_face\_landmarks.dat.bz2} to get the fixed number of keypoints for input face images. The SIFT features are calculated according to each keypoints with the public algorithm. 
Noting that we utilize the same hand-crafted feature extraction module in both the forgery detection classifier and the forgery type classifier.

Then, we aim to explore the multi-modality forgery clues in the frequency domain.
Motivated from related works \cite{chen2021local,frank2020leveraging}, we find there indeed exist noticeable different patterns scattered through the DCT spectrum (as shown in Fig. \ref{fig:failure-case}).
These visible artifacts are likely caused by upsampling operations in GAN models.
Further, we similarly utilize the ResNet model to extract discriminative deep network features.
It noted that we design two different deep network models with similar architecture in the forgery detection classifier and the forgery type classifier.
Here the ResNet$_{b}$  model is designed for the former forgery detection stage, and the ResNet$_{g}$ model is designed for the latter forgery type tracing stage in the image domain. 
In the frequency domain, the ResNet-DCT$_{b}$ model is only used in the first forgery detection stage.
When inputting nine local patches, the ResNet-DCT$_{b}$ model can capture nine deep network features and then cascade into the feature with the size of $9\times2048$.

To further effectively explore the dependencies between different image patches, we designed the AttenFusion Block to process mentioned extracted features. The self-attention mechanism and multi-head attention layer are brought here to grasp long-range relationship information of patch features, which is inspired by related work \cite{vaswani2017attention, zhao2021multi, zhao2022self, wang2022adt}. 
The multi-head attention layer can be deemed as an assembly of $n\_head$ self-attention, which can map each query and a set of key-value pairs. 
To be specific, the multi-attention layer will be weighted with $Q_{i}$, $K_{i}$, and $V_{i}$ for each head, which can be obtained from input feature vector $Fea$ and updated when training. 
The designed network utilizes these vectors to perform the self-attention calculation of each head, and then multiply a weight matrix $W$ after concatenating each head attention result to produce the output $Z$. 
The mentioned multi-head attention layer is formulated as:
\begin{align}
\label{eq3}
    Multi(Fea,W)=Concat(H_{1},...,H_{n\_head})W,
\end{align}
where $H_{i}=softmax(\frac{Q_{i}K_{i}^T}{\sqrt{d_k}})V_{i}$,
d$_{k}$ is the dimension of $K$ which is used for scaling to avoid gradient problems\cite{vaswani2017attention}. 
Following, the output will then be fed into the normalization layer, including residual connection and layer normalization:
\begin{equation} 
\label{eq5}
    Fea_i'=Norm(Fea_i+Multi(Fea_i,W)),
\end{equation}
where $Fea_i$ is the feature extracted from the local patches in a batch.
It is noted that the structure of the multi-head attention layer for hand-crafted features is different from that for deep network features. 
The former network aims to ensure each local spatial feature is calculated at different keypoints, and the latter aims to ensure each local feature is calculated in each image patch.
Following the Multi-layer perceptron (MLP) network is concatenated after the multi-head attention network to capture the forgery score.
Consequently, the output of the designed AttenFusion Block is formulated as:
\begin{equation} 
\label{eq6}
Score=MLP(Fea_1',...,Fea_N'),
\end{equation}
where $N$ is the number of image input features, which is calculated according to the number of patches and keypoints.

\subsection{Hierarchical Face Forgery Classifier}

Considering the property of mentioned multi-modality forgery faces, 
we design a hierarchical face forgery classifier consisting of authenticity feedback and adaptive refined weights to boost performance. It is because there exist class unbalanced distributions in data sets and it will bring adverse effects on forgery analysis accuracy. Different former works \cite{wang2022forgerynir} regard the task as a simple multi-class classification problem, the designed authenticity feedback strategy first distinguishes the authenticity and then recognizes the artifact types separately through the authenticity filter. The details of authenticity feedback are shown in Fig. \ref{fig:framework}. 
In the first forgery detection classifier stage, we separately train the frequency domain representation and image domain representation with the mentioned SIFT feature, the ResNet$_b$ feature, and the ResNet-DCT$_b$ feature. For further fusing different complementary discriminative information in image and frequency domains, we utilize the adaptive refined weighting to arrange learnable weights to obtain scores to determine the input face authenticity in an alternative way. Here we design three hyperparameters $w_{1,1}$, $w_{1,2}$, and $w_{1,3}$ to assign reasonable refined weights to these mentioned scores. It is noted that these three hyperparameters are learned alternatively through gradient descent until the objective loss converges. Thus, we can obtain different refined weights for these scores in different experimental settings. The refined authenticity score is formulated as:
\begin{equation}
\label{eq7}
S_{det}=\sum_{i=1}^3 \frac{exp({w_{1,i}}) \times S_{1,i}}{\sum_{j=1}^3 exp({w_{1,j}})},
\end{equation}
where $S_{1,1}$, $S_{1,2}$, $S_{1,3}$ are denoted as probability scores when inputting the ResNet-DCT$_{b}$ feature, the ResNet$_{b}$ feature, and the SIFT feature in both frequency and image domain through the designed AttenFusion block.

In the second forgery type classifier stage, we first utilize the former binary classifier as the authenticity filter to remove real faces in the training set to alleviate the class unbalanced problem. Here we only choose the ResNet$_g$ feature and the SIFT feature as the input features into the designed image domain classifier to capture probability scores. The reason is that the artifact types clues in the frequency domain are not distinct enough to recognize forgery types compared with the image domain. Following the same adaptive weighting strategy, we also design two hyperparameters $w_{2,1}$, $w_{2,2}$ to assign adaptive refined weights to these two scores.
\begin{equation}
\label{eq8}
S_{typ}=\sum^{2}_{i=1}\frac{exp({w_{2,i}}) \times S_{2,i}}{\sum_{j=1}^2 exp({w_{2,j}})},
\end{equation}
where $S_{2,1}$ and $S_{2,2}$ are the probability scores obtained by the SIFT feature and the ResNet$_{g}$ feature in the image domain through the designed AttenFusion blocks.

\subsection{Loss Function}
There exist several loss functions in our proposed algorithm in the training stage.
We design a simple patch-based Multi-layer Perceptron (MLP) in the designed AttenFusion Block for features extracted from each patch in order to obtain suitable local forgery clues. 
It noted that MLP is only used in the training process and will be removed when testing. 
In order to extract more discriminative forgery clues from individual face patches, the designed loss function mainly contains two types: loss from the patch-based model for each patch and loss from the hybrid domain representation model with patch fusion strategy.
The mentioned loss function can be formulated as:
\begin{equation}
\label{eq9}
Loss_{total}= \alpha*Loss_{pat}+(1-\alpha)*Loss_{fus},
\end{equation}
where $Loss_{pat}$ is the loss of extracting discriminative information from individual patches, and $Loss_{fus}$ is the loss of the AttenFusion Block in our proposed method.
The effect of parameter $\alpha$ is evaluated in the following experiments.

In the first forgery detection stage, we consider the problem as the typical binary classification.
Thus, the simple cross-entropy loss is used here to update parameters:
\begin{align}
Loss_{pat}=\sum^{N}_{i=1}\sum^{9}_{j=1}y_{ij} \log x_{ij}+(1-y_{ij}) \log (1-x_{ij}) \\
Loss_{fus} =\sum^{N}_{i=1}y_{i} \log x_{i}+(1-y_{i}) \log (1-x_{i}),
\end{align}
where $y_{ij}$ or $y_{i}$ is set to 1 if the original face image is fake, otherwise it is set to 0; $x_{i}$ denotes the predicted probability of input face images and $x_{ij}$ denotes that of divided patches from it.

In the second forgery type classification stage, we train these model parameters using the cross-entropy loss:
\begin{align}
Loss_{pat}= \sum_{i=1}^{N}\sum^{9}_{j=1}\sum_{m=1}^{M} y_{ij}^{m} \log \frac{e^{x_{ij}^{m}}}{\sum_{m=1}^{M}e^{x_{ij}^{m}}},  \\
Loss_{fus}= \sum_{i=1}^{N}\sum_{m=1}^{M} y_{i}^{m} \log \frac{e^{x_{i}^{m}}}{\sum_{m=1}^{M}e^{x_{i}^{m}}},
\end{align}
where M is the number of forgery face types, $y_{i}^{m}$ or $y_{ij}^{m}$ is set to 1 if the original face image belongs to class m, otherwise it is set to 0; $\frac{e^{x_{i}^{m}}}{\sum_{m=1}^{M}e^{x_{i}^{m}}}$ denotes the predicted probability belonging to class $m$ of the original face image and $\frac{e^{x_{ij}^{m}}}{\sum_{m=1}^{M}e^{x_{ij}^{m}}}$ denotes that of divided patches from it.

\textbf{Preprocessing.}
We first detect faces in raw images and then crop input images with existing public code. 
Noting that we don’t perform any resize operation in this stage, because it would bring a certain negative impact on the discrete cosine transform spectrum analysis. 
Besides, we divide the holistic face image into nine patches of size 128×128 for extracting the local face forgery discriminative information. These nine patches might overlap with each other, which contain four corner patches, four side patches and one center patch.
Following, each input patch is transformed into the frequency domain using the discrete cosine transform  with normalization:
$x^{*}_{i}=\frac{DCT(x_{i})-m}{\sqrt{var}}, $
where $x_{i}$ is the input raw image patch, and $x^{*}_{i}$ is the preprocessed image patch; $m$ and $var$ separately mean the mean and variance of input real images DCT spectrum.


\textbf{Network Architecture.}
For the deep convolution models for feature extraction, we reconstruct the classifier layers of the ResNet-50 network to get better performance. The reconstructed classifier layer contains two FC layers and one ReLU layer. 
The convolution layers of all ResNet models are initialized by the pre-trained on ImageNet dataset. 
All the networks in our proposed method are optimized with Adam optimizer. The learning rates for the $ResNet$ model, the $ResNet_{DCT}$ model are 1e-4,
while the learning rate for the designed AttenFusion Block is 1e-5.

Experiments on the ForgeryNIR dataset and the WildDeepfake dataset are all conducted on RTX 3090 GPU and pytorch platform. 
We detect input faces via open-source dlib \footnote{http://dlib.net/files/shape\_predictor\_68\_face\_landmarks.dat.bz2} and get 68 landmarks as the keypoints of the input face. 
In the following experiments, we utilized the mentioned adaptive refined weighting strategy to fuse different domain discriminative forgery clues, and the values are not fixed. In the WildDeepfake dataset, the optimal values of $w^1_1$, $w^1_2$, and $w^1_3$ are separately 1.36, 0.75, and 0.73. The optimal learnable weights are diverse in different settings.

\section{Experiments} \label{section:Experiments}
\begin{figure}
    \centering
    \includegraphics[width=0.5\textwidth]{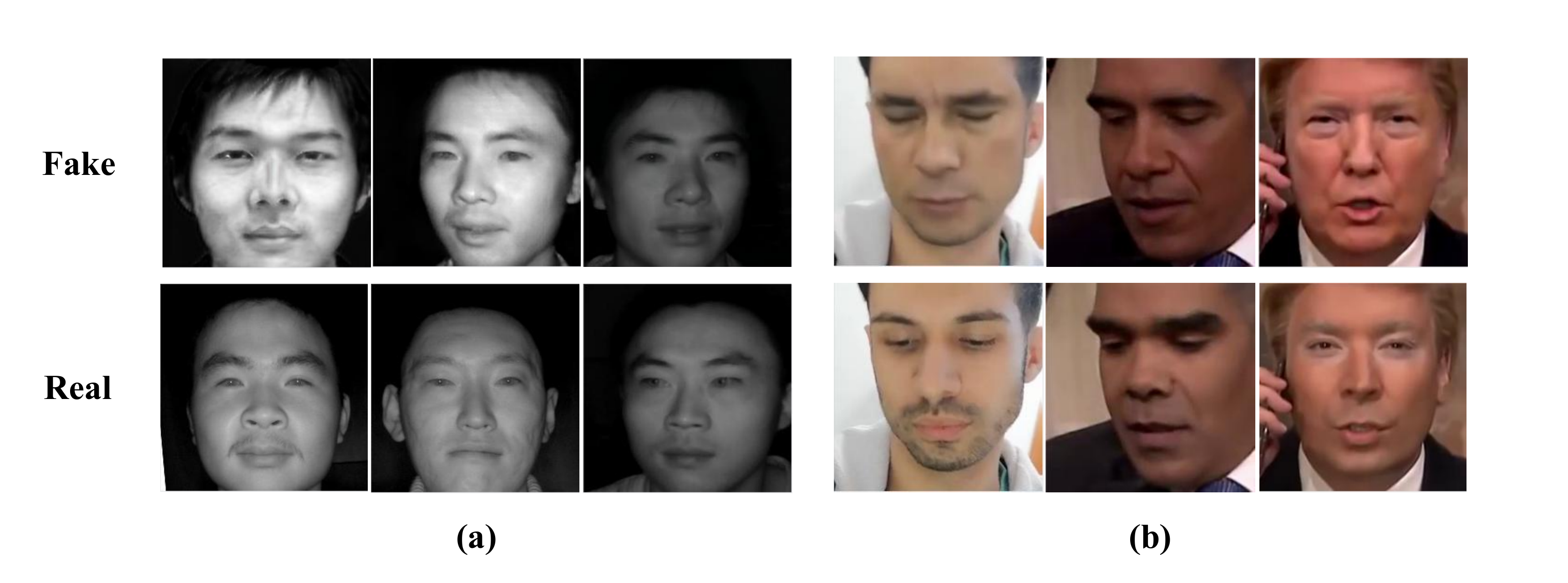}
    \caption{The samples of public multi-modality face forgery datasets: (a) ForgeryNIR Dataset; (b) WildDeepfake Dataset.}
    \label{fig:sample}
\end{figure}

In this section, we evaluate the performance of the proposed algorithm on three face datasets: the CASIA NIR-VIS 2.0 dataset\cite{li2013casia}, the ForgeryNIR dataset\cite{wang2022forgerynir}, and the WildDeepfake dataset\cite{zi2020wilddeepfake}, which represent two different modalities image (near-infrared (NIR) images and visible light-based (VIS) images). 
The samples of mentioned multi-modality face forgery detection are as shown in Fig. \ref{fig:sample}. 
We will give more details of experiment settings, and sufficient experimental results illustrate the effectiveness of the proposed algorithm.

\subsection{Datasets}
For detecting forgery clues and forgery tracing types in multi-modality face images, we conduct experiments on the CASIA NIR-VIS 2.0 dataset and the ForgeryNIR dataset. 
The CASIA NIR-VIS 2.0 dataset contains 17,580 real NIR images, and we select 10,000 images with real labels to compose the NIR-Real dataset. The ForgeryNIR dataset contains 240,000 forgery NIR images generated via multiple GAN techniques. The ForgeryNIR dataset has three different subsets, which is separately named as ForgeryNIR-std (images without any perturbation), ForgeryNIR-rand (images subject to only one perturbation), and ForgeryNIR-mix (images subject to more than one perturbation).
Moreover, this face forgery dataset also can be classified as single-epoch type (images generated by the same epoch models of the same GAN) or multi-epoch type (images generated by different epoch models of the same GAN).   

For detecting forgery clues of VIS face images, we conduct experiments on the public WildDeepfake dataset. 
The WildDeepfake database consists of 7,314 face sequences extracted from 707 deepfake videos entirely collected from the internet, making the dataset more challenging and closer to real-world scenarios. 
We conduct forgery detection experiments with face images provided by the WildDeepfake dataset which has been divided into training and testing sets, with a total of 1,014,436 images extracted from 707 videos. 
Noting that we do not conduct experiments of forgery tracing since there is no label of forgery types for the forgery image in the WildDeepfake dataset.

\begin{table*}[t]
\centering
\caption{Comparison of our five-classification experiment results on NIR datasets with the results of GANFingerprint\cite{yu2019attributing}, CNNDetection\cite{wang2020cnn}, GANDCTAnalysis\cite{frank2020leveraging}, Wavelet-Packet\cite{wolter2021wavelet}, Xception\cite{chollet2017xception}, and MLP\cite{wang2022forgerynir} provided by the ForgeryNIR dataset. We apply accuracy score(ACC) as the evaluation metrics}
\begin{tabular}{cccccccccc} 
\hline
 & Train & Test & GANFingerprint & CNNDetection & GANDCTAnalysis & \begin{tabular}[c]{@{}c@{}}Wavelet-\\ Packet\end{tabular} & Xception & MLP     
 & \begin{tabular}[c]{@{}c@{}}HFC-MFFD\\(Ours)\end{tabular}\\ \hline
                  &       & std  & $99.93$        & $100.00$        & $100.00$         & $99.64$        & $99.97$  & $46.98$  
                  & $\mathbf{100.00}$                       \\ 
                     & std   & rand & $56.93$        & $73.11$      & $39.23$        & $52.38$        & $56.14$  & $18.63$ 
                     &$\mathbf{93.20}$                     \\ 
                     &       & mix  & $59.56$        & $50.20$      & $30.43$        & $33.85$        & $46.31$  & $14.69$ 
                     & $\mathbf{94.29}$                         \\ \cline{2-10}
(1)                     & rand  & rand & $96.66$        & $98.25$      & $86.03$        & $87.87$        & $98.64$  & $26.70$ 
& $\mathbf{98.91}$                   \\
                     &       & mix  & $95.82$        & $92.04$      & $59.14$        & $69.51$        & $97.03$  & $29.18$ 
                     & $\mathbf{98.42}$                        \\ \cline{2-10}
                     & mix   & mix  & $98.14$        & $98.88$      & $70.94$        & $74.27$        & $98.90$  & $29.30$ 
                     &$\mathbf{99.61}$                        \\ \hline
                     &       & std  & $99.92$        & $99.99$      & $99.99$        & $98.03$        & $99.19$  & $43.33$ 
                     &$\mathbf{99.99}$                    \\ 
                     & std   & rand & $52.80$        & $67.54$      & $27.80$        & $61.74$        & $59.88$  & $23.51$ 
                     &$\mathbf{92.61}$                       \\  
                     &       & mix  & $52.33$        & $47.71$      & $20.58$        & $35.17$        & $51.51$  & $14.44$ 
                     &$\mathbf{89.43}$                      \\ \cline{2-10} 
(2)                     & rand  & rand & $93.42$        & $97.20$      & $82.66$        & $87.37$        & $94.37$  & $27.17$ 
&$\mathbf{98.89}$                   \\ 
                     &       & mix  & $92.22$        & $91.01$      & $56.32$        & $69.17$        & $94.02$  & $28.85$ 
                     & $\mathbf{97.37}$                        \\ \cline{2-10} 
                     & mix   & mix  & $94.76$        & $97.56$      & $64.82$        & $68.62$        & $96.15$  & $29.00$ 
                     &$\mathbf{99.53}$                     \\ \cline{1-10} 
\end{tabular}
\label{Table:five-classification-result}
\end{table*}

\subsection{Experiments on NIR Face Forgery Dataset}
To verify the effectiveness and robustness of the proposed algorithm on the NIR face dataset, we conduct sets of experiments on the ForgerNIR dataset and the NIR-Real dataset. Then we compare our proposed algorithm's results with the baseline results that the ForgeryNIR dataset provides. Firstly, we divide the ForgeryNIR dataset and the NIR-Real dataset into the training set, the validation set, and the testing set with a ratio of 7:1:2. Then, We conduct intra-dataset experiments and cross-dataset experiments. The difference between intra-dataset experiments and cross-dataset experiments is whether the number of perturbations is the same for the training and test sets. In these experiments, we used the training set to train the feature extraction models and the Forgery Classifier Block. The testing set will be then passed through the trained feature extraction model and the trained Forgery Classifier Block to obtain the possibility scores of the images. Finally, the classification scores are summed to get the final result.

The ForgeryNIR-mix under multi-epoch contains forgery images generated by models of different periods subject to multiple perturbations, which are closer to realistic scenarios. Therefore, using the ForgeryNIR-mix under multi-epoch as the testing set is more challenging, especially the intra-dataset experiments with ForgeryNIR-mix under multi-epoch as the training set and the cross-dataset experiments with ForgeryNIR-std under multi-epoch as the training set are the most challenging experiments. Therefore, we take these three experimental results as the main indicators to evaluate the performance of the algorithm. 

The comparison results are shown in Table \ref{Table:five-classification-result}, although the CNNDetection method has good effectiveness (performance in experiments on intra-dataset experiments) and robustness (performance in cross-datasets experiments) among the six baseline results, our method can achieve much better performance. To be specific, our proposed method outperforms the baseline results in almost all experiments except the intra-dataset experiment with ForgeryNIR-std in the multi-epoch as the training set. For the intra-dataset experiment with ForgeryNIR-mix in the multi-epoch as the training set, our method can achieve an accuracy of 99.53\%, which is 1.97\% higher than the highest accuracy in the baseline results, proving the effectiveness of our proposed method. 
And for the cross-dataset experiments with ForgeryNIR-std in the multi-epoch as the training set, most of the methods shown in Table \ref{Table:five-classification-result} have a poor performance, while our method can achieve an accuracy of 92.61\% when the testing set is ForgeryNIR-rand and 89.43\% when the testing set is ForgeryNIR-mix, which means our proposed method can effectively explore the forgery clues in ForgeryNIR-mix type dataset. 
Surprisingly, the results also prove that our proposed method could achieve a satisfactory robustness performance in cross-dataset experiments.  

\subsection{Experiments on VIS Face Forgery Datasets}
In this subsection, we conduct experiments on the WildDeepfake dataset to further verify the effectiveness of our proposed algorithm on the VIS modality and then compare it with related methods.
The WildDeepfake dataset has an unbalanced amount of data under the real and fake labels (the training set has 381,875 images under the real label and 632,561 images under the fake label in the image level), which further increases the difficulty of classification. 
Here we use the same protocol with \cite{zi2020wilddeepfake}
for fair comparisons.

Due to the absence of forgery-type labels, we only verify the SOTA performance of our proposed algorithm in the first stage on the VIS modality. The experimental results are shown in Table \ref{Table:the result in the WildDeepfake dataset}. These experimental results demonstrate the effectiveness of our proposed method. Our proposed method can achieve an accuracy of 85.59\%, which is 2.27\% higher than DAM\cite{zhao2021multi}, and 2.34\% higher than RECCE\cite{cao2022end}.
F3-Net\cite{qian2020thinking} can effectively explore forgery clues in the frequency domain, and achieve an accuracy of 80.66\%. Our proposed method achieves the best performance because the HFC-MFFD could explore robust discriminative forgery clues effectively.

\begin{table}[ht] 
\centering
\caption{Comparison of the proposed method experiment results on the WildDeepfake dataset with the results provided by the WildDeepfake dataset.}
\begin{tabular}{cc}
\hline
Methods    & ACC(\%) \\ \hline
AlexNet\cite{krizhevsky2012imagenet}    & 60.37  \\
VGG-16\cite{simonyan2014very}     & 60.92  \\
ResNetV2-50\cite{he2016deep} & 63.99 \\
Inception-v2\cite{szegedy2016rethinking} &58.73 \\
MesoNet-inception\cite{afchar2018mesonet} & 66.03 \\
ADDNet-2D \cite{zi2020wilddeepfake}  & 76.25    \\
XceptionNet\cite{chollet2017xception}  & 79.99   \\ 
Capsule\cite{nguyen2019capsule}     & 78.68   \\
ADD-Xception\cite{khormali2021add} & 79.23  \\
RFM\cite{wang2021representative}         & 77.38  \\
F3-Net\cite{qian2020thinking}    & 80.66   \\
DAM\cite{zhao2021multi}         & 83.32  \\
RECCE\cite{cao2022end}      & 83.25   \\
\textbf{HFC-MFFD (Ours)}      & \textbf{85.59} \\ \hline

\end{tabular}
\label{Table:the result in the WildDeepfake dataset}
\end{table}

\subsection{Algorithm Analysis}
In this subsection, we further analyze our proposed algorithm. We conduct a series of ablation studies on the proposed algorithm to evaluate its effectiveness and robustness ability.
To be specific, we compared the experimental results after removing its key components.
Noting that we choose ResNet and SIFT algorithms to represent mentioned deep network feature and hand-crafted feature in the following experiments.
There exist three kinds of extracted features: the ResNet feature, the SIFT feature and the ResNet-DCT feature. The ResNet feature and the ResNet-DCT are different in the first and second stages because they are extracted by different deep network models. 
Although the ResNet-DCT feature is not used in the second stage of our proposed algorithm, we still train a ResNet-DCT model and then extract the ResNet-DCT feature in the second stage for comparison.
For the convenience of comparison, we set mentioned score weights as 1 in the analysis experiments.
It is noted that Each subset is divided into the training set, the validation set and the testing set with a ratio of 7:1:2. 

\begin{table}[t]       
\centering
\caption{Ablation study on the first stage of our proposed algorithm to evaluate the effectiveness of different feature extraction when ForgeryNIR-mix under multi-epoch selected as the training set and testing set.}
\begin{tabular}{cccc}
\hline
\multicolumn{2}{c}{Image domain} & Frequency domain & \multirow{2}{*}{ACC(\%)} \\ \cline{1-3}
 deep feature         & hand-crafted         & deep feature        &                          \\ \hline
                      \checkmark                 &      -        &      -            & $100.00$                   \\ 
                                -        & \checkmark            &    -              & $98.56$                    \\
                          -     &     -         & \checkmark                & $100.00$                      \\  
                      \checkmark                 & \checkmark            &           -       & $99.56$                    \\ 
                      \checkmark                 &    -          & \checkmark                & $100.00$                    \\ 
                       -               & \checkmark            & \checkmark                & $99.64$                      \\ 
                     \checkmark                 & \checkmark            & \checkmark                & $99.94$                      \\ \hline
\end{tabular}
\label{Table:ablation study-effectiveness-first stage}
\end{table}

\begin{table}[t]    
\centering
\caption{Ablation study on the second stage of our proposed algorithm to evaluate the effectiveness of different feature extraction when ForgeryNIR-mix under multi-epoch selected as the training set and testing set.}
\begin{tabular}{cccc}
\hline
 \multicolumn{2}{c}{Image domain} & Frequency domain & \multirow{2}{*}{ACC(\%)} \\ \cline{1-3}
 deep feature         & hand-crafted         & deep feature        &                          \\ \hline
                           \checkmark                 &   -           & -                 & $98.51$                  \\ 
                                       -      & \checkmark            &  -                & $94.01$                   \\ 
                                   -         &         -     & \checkmark                & $94.55$                   \\ 
                           \checkmark                 & \checkmark            &    -              & $98.54$                   \\ 
                          \checkmark                 &    -          & \checkmark                & $98.65$                   \\ 
                                      -       & \checkmark            & \checkmark                & $97.73$                   \\ 
                           \checkmark                 & \checkmark            & \checkmark                & $\mathbf{98.94}$    \\ \hline             
\end{tabular}
\label{Table:ablation study-effectiveness-second stage}
\end{table}

\subsubsection{The comparison of different feature extraction strategies} 
The challenging ForgeryNIR-mix setting contains images generated by models of different periods with different disturbances.
Therefore, we use ForgeryNIR-mix under multi-epoch as the dataset in this ablation study. 
Firstly, we evaluate the effectiveness of the proposed feature extraction strategy in the first stage of our proposed algorithm. 
As shown in Table \ref{Table:ablation study-effectiveness-first stage}, only using the ResNet-DCT$_{b}$ feature can achieve an accuracy of 100\%, which indicates that information in the frequency domain can help perform forgery detection task. 
It also proves that binary forgery classification seems an easily accessible task in the mentioned dataset.

In addition, we conduct an extra ablation study to evaluate the feature extraction strategy effectiveness in the second stage of our proposed algorithm. 
As shown in Table \ref{Table:ablation study-effectiveness-second stage}, we perform the multiple forgery type tracing experiment, which uses fake images selected in the former stage. 
This experimental result demonstrates that fusing both the deep feature and hand-crafted feature in image and frequency domains could obtain a comparatively good performance (98.54\%) for the forgery tracing task, which is 0.03\% and 4.53\% higher than only using the deep feature and hand-crafted feature separately. 
It is because the proposed local hybrid domain forgery representation could effectively explore robust discriminative forgery clues to enhance detection performance.

\begin{table}[t]    
\centering
\caption{Ablation study on the first stage of our proposed algorithm to evaluate the robustness of different feature extraction when ForgeryNIR-std under multi-epoch selected as the training set, and ForgeryNIR-mix under multi-epoch selected as testing set}
\begin{tabular}{ccccc}
\hline
Type                 & \multicolumn{2}{c}{Image domain} & Frequency domain & \multirow{2}{*}{ACC(\%)} \\ \cline{2-4}
\multicolumn{1}{l}{} & deep feature         & hand-crafted         & deep feature        &                          \\ \hline
                     & \checkmark                 & -            & -                & $98.27$                    \\ 
                     & -                 & \checkmark            & -                & $98.17$                    \\
                     & -                 & -            & \checkmark                & $91.23$                    \\
mix                  & \checkmark                 & \checkmark            & -                & $98.84$                    \\
                     & \checkmark                 & -            & \checkmark                & $99.55$                    \\
                     & -                 & \checkmark            & \checkmark                & $99.29$                    \\ 
                     & \checkmark                 & \checkmark            & \checkmark                & $\mathbf{99.75}$                    \\ \hline
\end{tabular}
\label{Table:ablation study-robustness-first stage}
\end{table}

\begin{table}[t]    
\centering
\caption{Ablation study on the second stage of our proposed algorithm to evaluate the robustness of different feature extraction when ForgeryNIR-std under multi-epoch selected as the training set, and ForgeryNIR-mix under multi-epoch selected as the testing set}
\begin{tabular}{ccccc}
\hline
Type                 & \multicolumn{2}{c}{Image domain} & Frequency domain & \multirow{2}{*}{ACC(\%)} \\ \cline{2-4}
\multicolumn{1}{l}{} & deep feature         & hand-crafted         & deep feature        &                      \\ \hline
                     & \checkmark                 & -            & -                & $55.81$                \\ 
                     & -                 & \checkmark            & -                & $85.05$              \\
                     & -                 & -            & \checkmark                & $35.18$               \\
mix                  & \checkmark                 & \checkmark            & -                & $\mathbf{85.53}$              \\
                     & \checkmark                 & -            & \checkmark                & $47.49$                 \\
                     & -                 & \checkmark            & \checkmark                & $75.68$               \\ 
                     & \checkmark                 & \checkmark            & \checkmark                & $77.03$                 \\ \hline
\end{tabular}
\label{Table:ablation study-robustness-second stage}
\end{table}

\subsubsection{The robustness of different feature extraction strategies} 
Similar to \cite{wang2022forgerynir}, we choose ForgeryNIR-std under multi-epoch as the training set and ForgeryNIR-mix under multi-epoch as the testing set for algorithm robustness evaluation. 
Firstly, we evaluate the robustness of the feature extraction in the first stage of our proposed algorithm. 
It noted that the ResNet$_{b}$ feature and SIFT feature are represented typical deep features and hand-crafted features in the image domain, and the ResNet-DCT$_{b}$ feature is represented typical deep feature in the frequency domain.
As shown in Table \ref{Table:ablation study-robustness-first stage}, fusing the ResNet$_{b}$ feature, the SIFT feature and the ResNet-DCT$_{b}$ feature can achieve 99.75\% in the mentioned protocol, which is higher than other feature fusion strategies. 
It is noted that adding the ResNet-DCT feature can improve 0.91\% compared to without it, which proves that the deep feature in the frequency domain can boost forgery detection performance.

Additionally, we conduct another ablation study to evaluate the feature extraction strategies' robustness in the second stage of our proposed algorithm. 
As shown in Table \ref{Table:ablation study-robustness-second stage}, fusing both the deep feature and hand-crafted feature only in the image domain can achieve the best performance in the cross-dataset experiments.
It is noted that the accuracies of only utilizing hand-crafted features in the image domain are 85.05\%, which is higher than other individual feature extraction strategies. 
It is because the mentioned hand-crafted features in the image domain can robustly identify objects even among clutter and under partial occlusion. These experimental results indicate that features in the frequency domain are not helpful to improve performance in the forgery tracing task under disturbance. Because these artifact type clues in the frequency domain are not distinguishable enough to recognize forgery types for multi-modality faces. According to this phenomenon, the designed authenticity feedback strategy explores both frequency domain and image domain forgery clues in the first forgery detection stage and only explores image domain forgery clues in the second forgery type classification stage.

\subsubsection{The effect of parameter patch size}
It is important to find a suitable patch size in the proposed method because small-size patches would lose inherent discriminative information and large-size patches may contain redundant information. Here we evaluate the effect of parameter patch size on the WildDeepfake dataset. As shown in Table \ref{Table:Evaluation of the hyperparameter patch size on the WildDeepfake dataset}, we evaluate the effect of parameter patch size from a set of $\{32, 64, 128, 144, 224\}$. We find that when the parameter patch size is set as 128, the forgery detection accuracy becomes better. Thus, we divide the input faces into several $128\times128$ patches in the following experiments.

\begin{table}[t]
\centering
\caption{Evaluation of the hyperparameter patch size on the WildDeepfake dataset}
\begin{tabular}{@{}cccc@{}}
\hline

Patch Size & Patch Num & Acc(\%) \\ \hline
$32\times32$      & $81 (9\times9)$   & $79.44$   \\
$64\times64$      & $25 (5\times5)$   & $80.83$   \\
$128\times128$    & $9 (3\times3)$    & $83.17$   \\
$144\times144$    & $9 (3\times3)$    & $81.72$   \\
$224\times224$    & $1 (1\times1)$    & $81.73$   \\ \hline

\end{tabular}
\label{Table:Evaluation of the hyperparameter patch size on the WildDeepfake dataset}
\end{table}

\subsubsection{The effect of parameter $\alpha$}
\begin{figure} 
    \centering
    \includegraphics[width=0.40\textwidth]{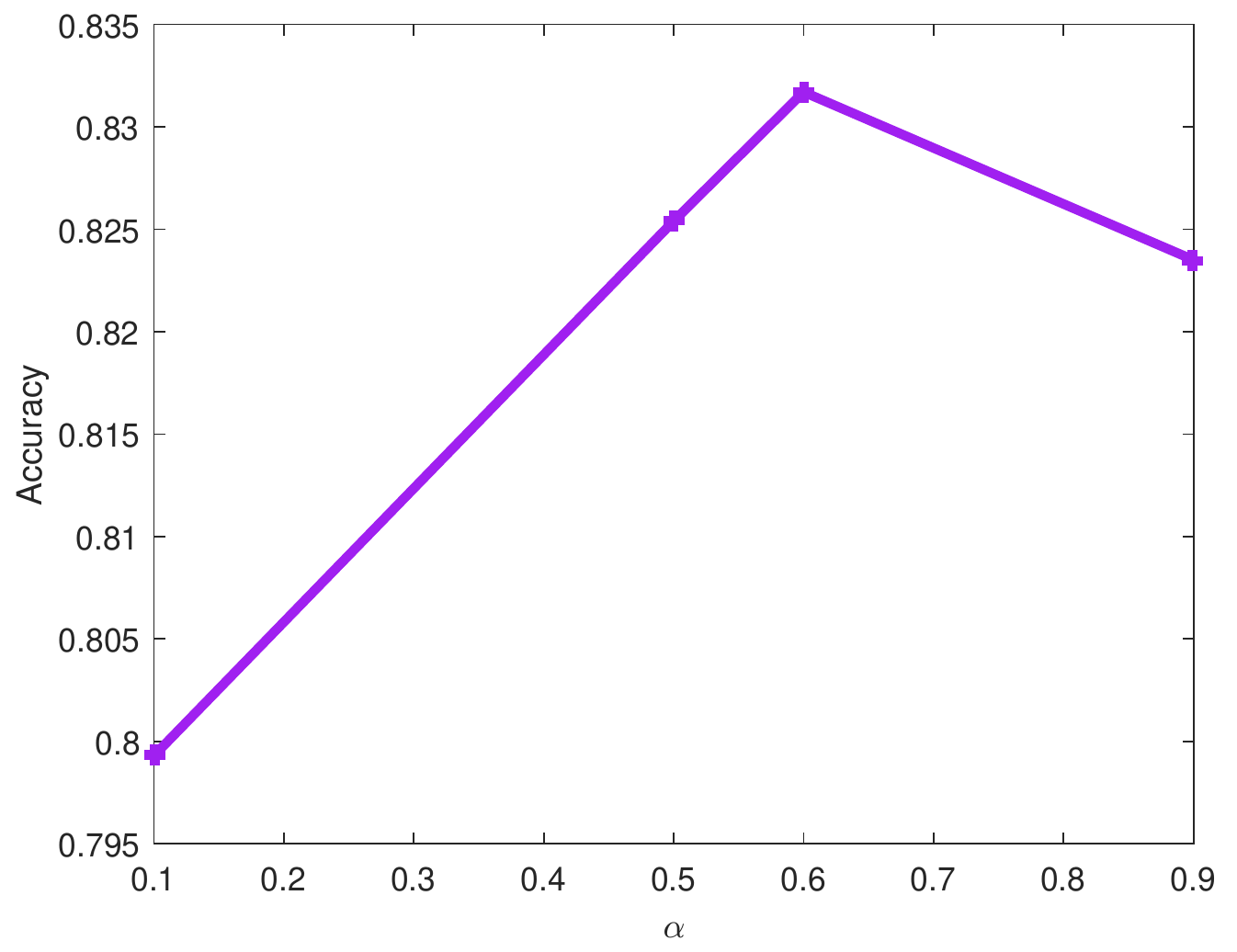}
    \caption{The effect of the hyperparameter $\alpha$ on the WildDeepfake dataset}
    \label{fig:Evaluation of the hyperparameter alpha on the WildDeepfake dataset}
\end{figure}

As mentioned in the Algorithm Section, parameter $\alpha$ balances the effect of patch-based loss term and hybrid domain representation loss term. Here we evaluate the effect of parameter $\alpha$ on the WildDeepfake dataset. As shown in Fig. \ref{fig:Evaluation of the hyperparameter alpha on the WildDeepfake dataset}, we evaluate the effect of parameter $\alpha$ from a set of $\{0.1, 0.5, 0.6, 0.9\}$. We find that when the parameter $\alpha$ achieves approximately 0.6, the performance achieves the best. Thus, we set the parameter $\alpha$ as 0.6 in the following experiments.

\subsubsection{Cross-domain evaluations}
As shown in Table \ref{Table: Evaluation of Generalizability of our proposed method}, in order to evaluate the cross-modality ability of our proposed method, we conduct the following cross-domain experiments. The ForgeryNIR dataset and WildDeepfake dataset are separately selected as representative VIS and NIR domain images here. The ForgeryNIR-mix in multiple epoch datasets is selected as the training set and the WildDeepfake dataset is selected as the testing set, and then swap the training set and testing set. Experimental results prove our proposed method only could achieve satisfactory performance in intra-domain evaluation, but achieve unsatisfactory performance in cross-domain evaluation. It is because there still exists a large domain gap between multi-modality images, which inspire more researcher to focus on multi-modality forgery generalization. 

\begin{table}[t] 
\centering
\caption{Evaluation of Generalizability of our proposed method where the ForgeryNIR dataset and the WildDeepfake dataset are relatively representative of NIR and VIS modality}
\begin{tabular}{@{}ccc@{}}
\hline
Train Set                                & Test Set                  & Acc(\%) \\ \hline
\multirow{2}{*}{ForgeryNIR} & ForgeryNIR & $100.00$   \\
                                       & WildDeepfake          & $63.57$   \\ \hline
\multirow{2}{*}{WildDeepfake}          & WildDeepfake          & $85.06$   \\
                                       & ForgeryNIR & $28.61$   \\ \hline
\end{tabular}
\label{Table: Evaluation of Generalizability of our proposed method}
\end{table}

\subsubsection{Failure Case}
\begin{figure}
    \centering
    \includegraphics[width=0.5\textwidth]{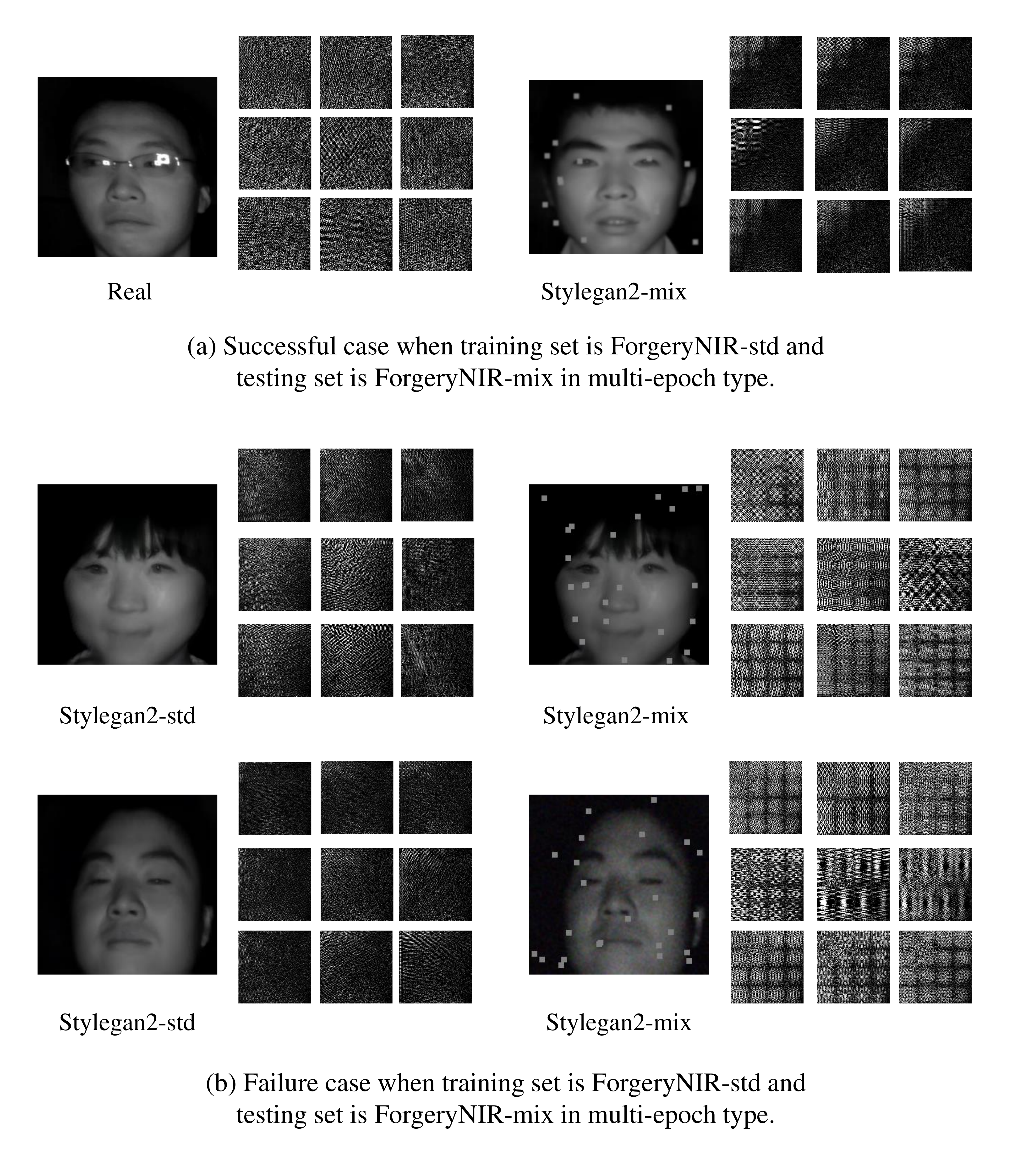}
    \caption{Examples of successful classification and misclassification in the first stage when the training set is ForgeryNIR-std and the testing set is ForgeryNIR-mix.}
    \label{fig:failure-case}
\end{figure}

\begin{figure}
    \centering
    \includegraphics[width=0.5\textwidth]{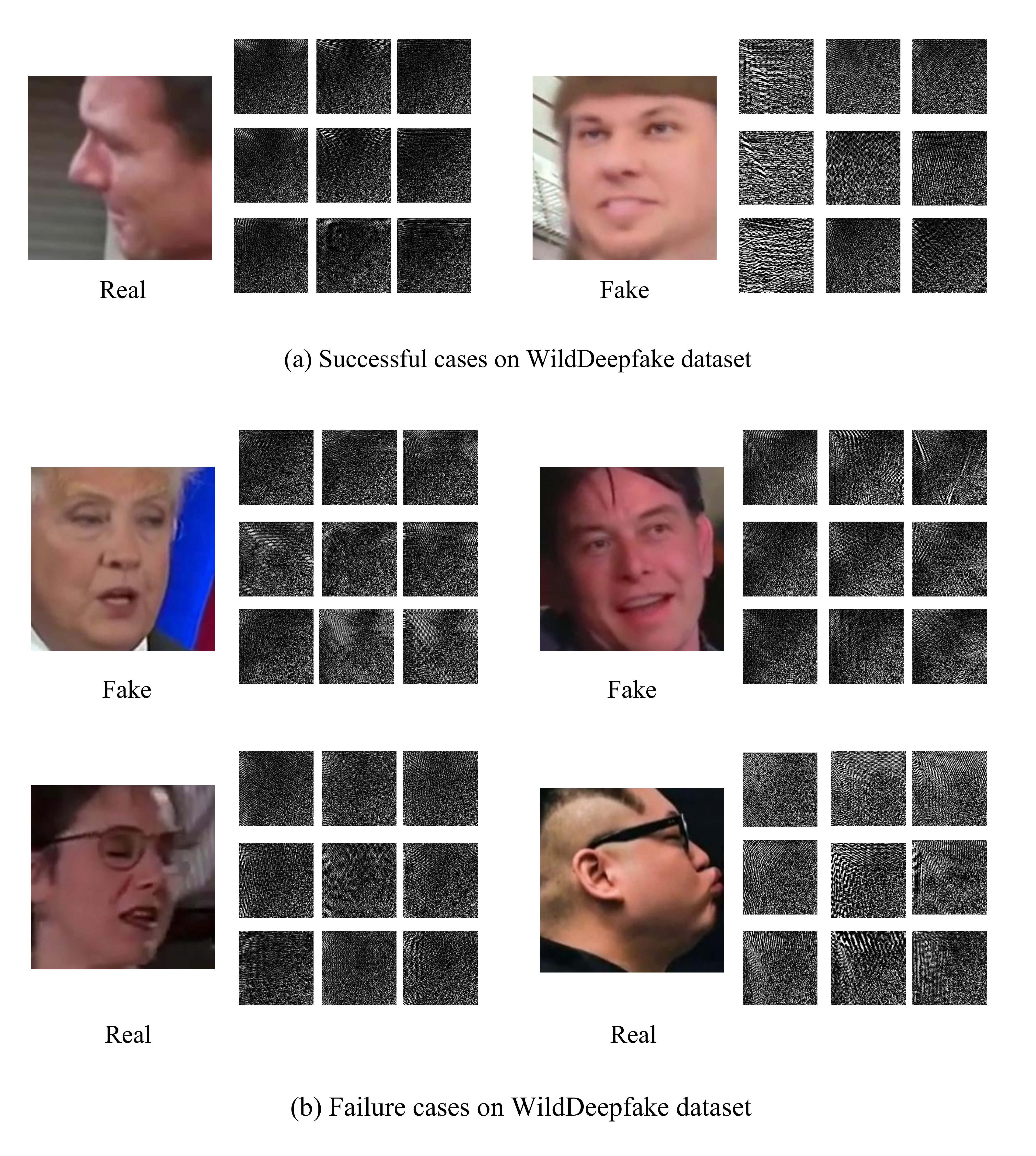}
    \caption{Examples of successful classification and misclassification in the forgery detection stage evaluated on the WildDeepfake dataset.}
    \label{fig:failure-case-wilddeepfake}
\end{figure}

Here we show several failure cases on the ForgeryNIR database with our proposed HFC-MFFD in Fig. \ref{fig:failure-case}. 
The failure cases show misclassified images in the first stage of the cross-dataset experiment, where the training set is ForgeryNIR-std and the testing set is ForgeryNIR-mix. 
To investigate the proposed algorithm clearly, we compared them with corresponding face images and the processed DCT spectrum.
In the first line, we show the real image and its nine patches' DCT spectrum to help analyze the difference between real and fake images. 
Additionally, we also show the correctly classified forgery image and the corresponding nine patches' DCT spectrum to help analyze failure cases.
Due to the normalization processing of the real image, the high-frequency components show a scattered state.
As shown in the DCT spectrum of the forgery image under ForgeryNIR-std, the low-frequency components are concentrated in the upper left corner, which also happened to the successfully classified forgery image under ForgeryNIR-mix. 
\textit{We assume that similar frequency properties make them closer to each other in the frequency domain.}
For these failure cases, the DCT spectrum of forgery images under ForgeryNIR-mix exists obvious grid-like artifacts likely due to the block-based distortion processing \cite{frank2020leveraging}.
These specific frequency properties seem closer to the real image in the first line, which could cause forgery detection failure.
A similar experiment analysis is conducted on WildDeepfake dataset as shown in Fig. \ref{fig:failure-case-wilddeepfake}. Different from the ForgeryNIR dataset, it is more difficult to explore the specific pattern scattered through the DCT spectrum for forgery faces, because of lacking the block-based distortion processing. However, we can still find there indeed exists a difference between real and fake faces, where the low-frequency components are relatively sparse and partly concentrated in the DCT spectrum of real images. We think the interesting visualization analysis would inspire more researchers to explore robust forgery clues in the real world.

\section{CONCLUSION} \label{section:Conclusion}
The paper proposes a novel hierarchical forgery classifier for the multi-modality face forgery detection task.
The proposed HFC-MFFD can effectively integrate complementary discriminative forgery clues in the score level and distinguish the authenticity and forgery types in order.
Considering the specific properties of forgery images, the local hybrid domain forgery representation is designed to explore robust discriminative features to boost detection performance.
Experimental results on several typical multi-modality forgery face datasets demonstrate the superior performance
of the proposed method compared with SOTA methods.
In the future, we would evaluate the proposed HFC-MFFD performance on more multi-modality forgery datasets and explore the cross-dataset robustness to mimic real-world scenarios.

\bibliographystyle{IEEEtran}
\bibliography{reference}

\vfill

\end{document}